%% file: main.tex
\documentclass[10pt,twocolumn,letterpaper]{article}
\usepackage{cvpr}
\usepackage{times}
\usepackage{epsfig}
\usepackage[dvipsnames]{xcolor}
\usepackage{graphicx}
\usepackage{amsmath}
\usepackage{amssymb}
\usepackage{amsthm}
\usepackage{bbm}
\usepackage{subfig}
\usepackage{caption}
\usepackage{booktabs} 
\usepackage{algorithm}
\usepackage{algorithmic}
\usepackage{multirow}
\usepackage{tabularx}
\usepackage{float}

\usepackage{svg}
\usepackage{times}
\usepackage{epsfig}
\usepackage{graphicx}
\usepackage{amsmath}
\usepackage{amssymb}

\usepackage{booktabs}
\usepackage{tabulary}

\usepackage[british,english,american]{babel}
\usepackage{soul}
\usepackage{xspace}\xspace
\usepackage{adjustbox}
\usepackage{array}

\usepackage{dblfloatfix}
\usepackage{transparent}

\usepackage{algorithm}
\usepackage{listings}
\usepackage{indentfirst}
\usepackage[bottom]{footmisc}

\cvprfinalcopy 

\def\cvprPaperID{5114} 

\ifcvprfinal\fi
\begin{document}

\title{Scale Calibrated Training: Improving Generalization of Deep Networks via Scale-Specific Normalization\thanks{Preprint}}

\author{Zhuoran Yu\thanks{Work done during internship at SenseTime}\\
Georgia institute of Technology\\
{\tt\small zhuoranyu@gatech.edu}
\and
Aojun Zhou\\
SenseTime Research\\
{\tt\small zhouaojun@sensetime.com}
\and 
Yukun Ma\\
SenseTime Research\\
{\tt\small mayukun@sensetime.com}
\and 
Yudian Li\\
SenseTime Research\\
{\tt\small liyudian@sensetime.com}
\and 
Xiaohan Zhang\\
SenseTime Research\\
{\tt\small zhangxiaohan@sensetime.com}
\and 
Ping Luo\\
The Univesity of Hong Kong\\
{\tt\small pluo@cs.hku.hk}
}

\maketitle


\begin{abstract}
   Standard convolutional neural networks(CNNs) require consistent image resolutions in both training and testing phase. However, in practice, testing with smaller image sizes is necessary for fast inference. We show that trivially evaluating low-resolution images on networks trained with high-resolution images results in a catastrophic accuracy drop in standard CNN architectures. We propose a novel training regime called \textit{Scale calibrated Training(SCT)} which allows networks to learn from various scales of input simultaneously. By taking advantages of SCT, single network can provide decent accuracy at test time in response to multiple test scales. In our analysis, we surprisingly find that 
  vanilla batch normalization can lead to sub-optimal performance in SCT. Therefore, a novel normalization scheme called \textit{Scale-Specific Batch Normalization} is equipped to SCT in replacement of batch normalization. Experiment results show that SCT improves accuracy of single Resnet-50 on ImageNet by 1.7\% and 11.5\% accuracy when testing on image sizes of 224 and 128 respectively.
\end{abstract}

\input{introduction}

\input{related}
\input{method}
\input{experiments}

\input{conclusion}



{\small
\bibliographystyle{ieee_fullname}
\bibliography{main}
}

\end{document}

%% file: introduction.tex
\section{Introduction}
Image classification, one of the most important fundamental task in computer vision, has gained a remarkable success with the development of convolutional neural networks(CNNs)\cite{krizhevsky2012imagenet, simonyan2014very, szegedy2015going, he2016deep, howard2017mobilenets, tan2019efficientnet}. The advent of Batch Normalization(BN)\cite{ioffe2015batch} also serves as an effective component of deep convolutional neural networks not only improving network accuracy but also allowing faster convergence of the training process.  

The standard regime of using CNN for classification involves network training and inference. Typically, images used in these two phases are in the same sizes. However, in practice, sizes of testing images may vary when the tradeoff of "accuracy vs speed"\cite{tan2019efficientnet} is exploited. Specifically, one may want to use smaller images in testing for faster inference time; on the other hand, when longer inference time is acceptable, large images are preferred for accuracy ensuring. 

\begin{figure}[H]
\centering
\vspace{-0.3cm} 
\includegraphics[width=0.5\textwidth]{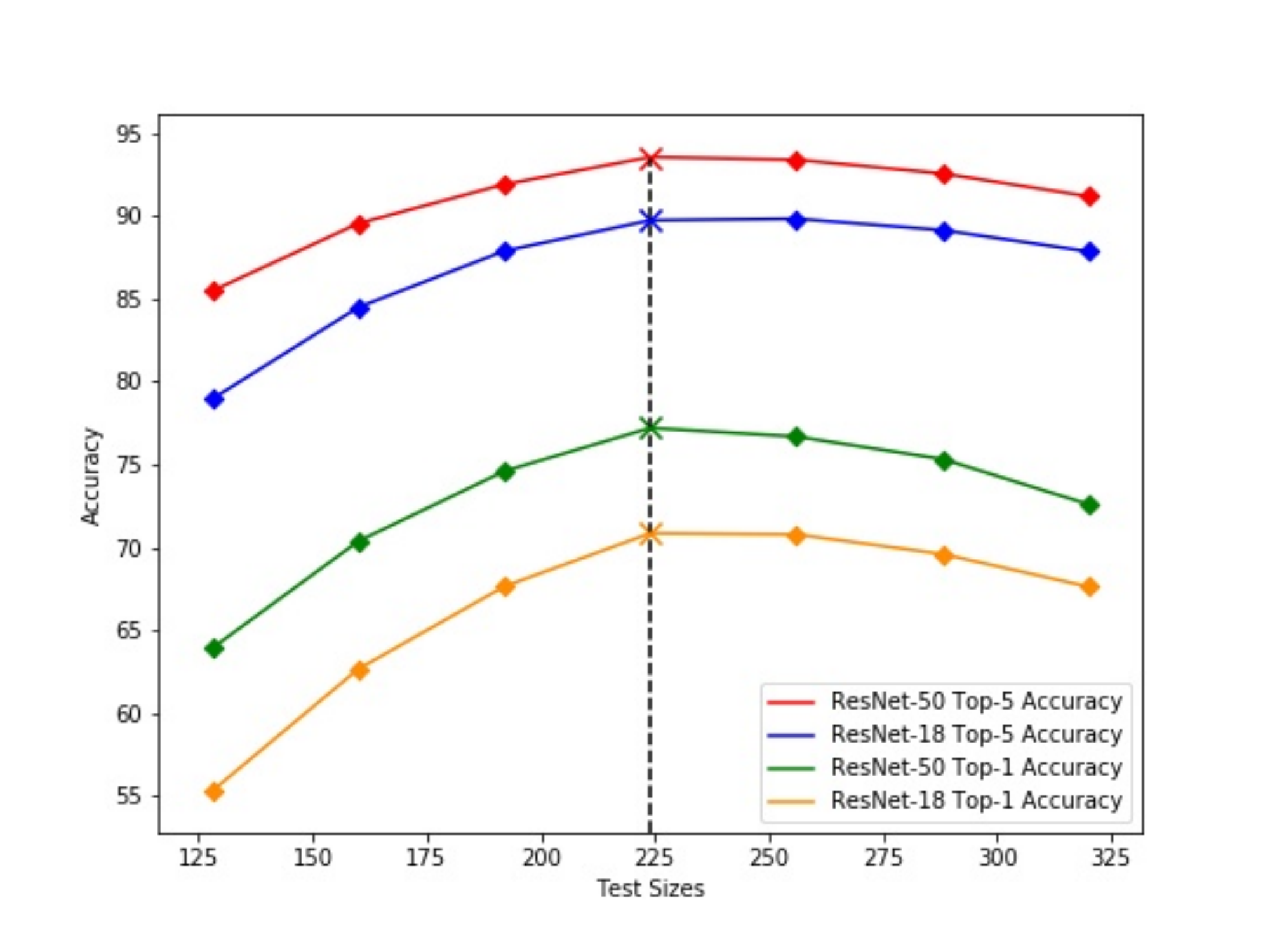}
\label{fig: intro}
\vspace{-5mm}
\caption{The above figure presents validation results of ResNet-18 and ResNet-50 trained on image sizes $224 \times 224$ and tested on various sizes.}
\label{fig: de}
\end{figure}

However, scale invariance is not perfectly guaranteed by standard CNNs due to non-linear downsampling and activations. Discrepancies between sizes of training and testing images naturally lead to accuracy decline\cite{touvron2019fixing}. The fact is illustrated in Figure \ref{fig: de}, where regular Resnets\cite{he2016deep} are trained on ImageNet\cite{deng2009imagenet} with input size of $224 \times 224$ and tested with various sizes from $320 \times 320$ to $128 \times 128$. It can be inferred from the figure that resolution inconsistency between training and testing causes catastrophic accuracy drop. 

Training with test resolution does not completely solve the problem. Low resolution images contain less semantic information which consequently lowers the upper bound of accuracy that networks can achieve. Moreover, it also introduces extra storage and management cost to maintain different models trained with different image sizes. If there is a demand of evaluating images with K different sizes, this trivial solution would factor the storage cost by a factor of K and introduce additional effort to manage all these models. 

In this paper, we describe a novel training regime called \textit{Scale Calibrated Training(SCT)} which improves performance of convolutional neural networks no matter what sizes of testing images are given. In general, SCT is composed by three phases: sampling phase, forwarding phase, and calibrating phase. During the sampling phase, a batch of images is sampled with several distinct spatial sizes and these samples are fed into convolutional layers which is known as the forward phase. Finally, in the calibrating phase, features extracted from samples at different resolution are passed to an calibrating layer where features are spatially calibrated to create fix-sized feature vectors for classification. 

We further identify the fact that vanilla batch normalization in response to various spatial sizes may lead to sub-optimal performance of deep convolutional neural networks. As we show in this paper, distribution discrepancy between different scales is exacerbated as the network goes deeper, which violates the assumption of batch normalization. To tackle this drawback of batch normalization,we propose an effective normalization process in SCT called \textit{Scale-Specific Normalization} which is particularly useful in the scenario that sizes of input image can vary. 

We summarize the contribution of this paper as follows:
\begin{enumerate}
    \item We illustrate the accuracy decline when testing with different image resolutions other than tge training resolution and propose \textit{Scale Calibrated Training} to tackle this problem. SCT steadily improves accuracy of a \textbf{single} network with various test resolution, especially for those are lower than the training resolution. 
    \item We identify the discrepancy between distribution of internal activation of networks with different input scales and demonstrate it deleterious effect on the performance of batch normalization.
    \item We propose a novel normalization module called \textit{Scale-Specific Normalization} which is particularly beneficial when inputs of batch normalization have scale variance.
\begin{figure*}[htp!]
\centering
\vspace*{-0.5in}
\includegraphics[scale=0.3]{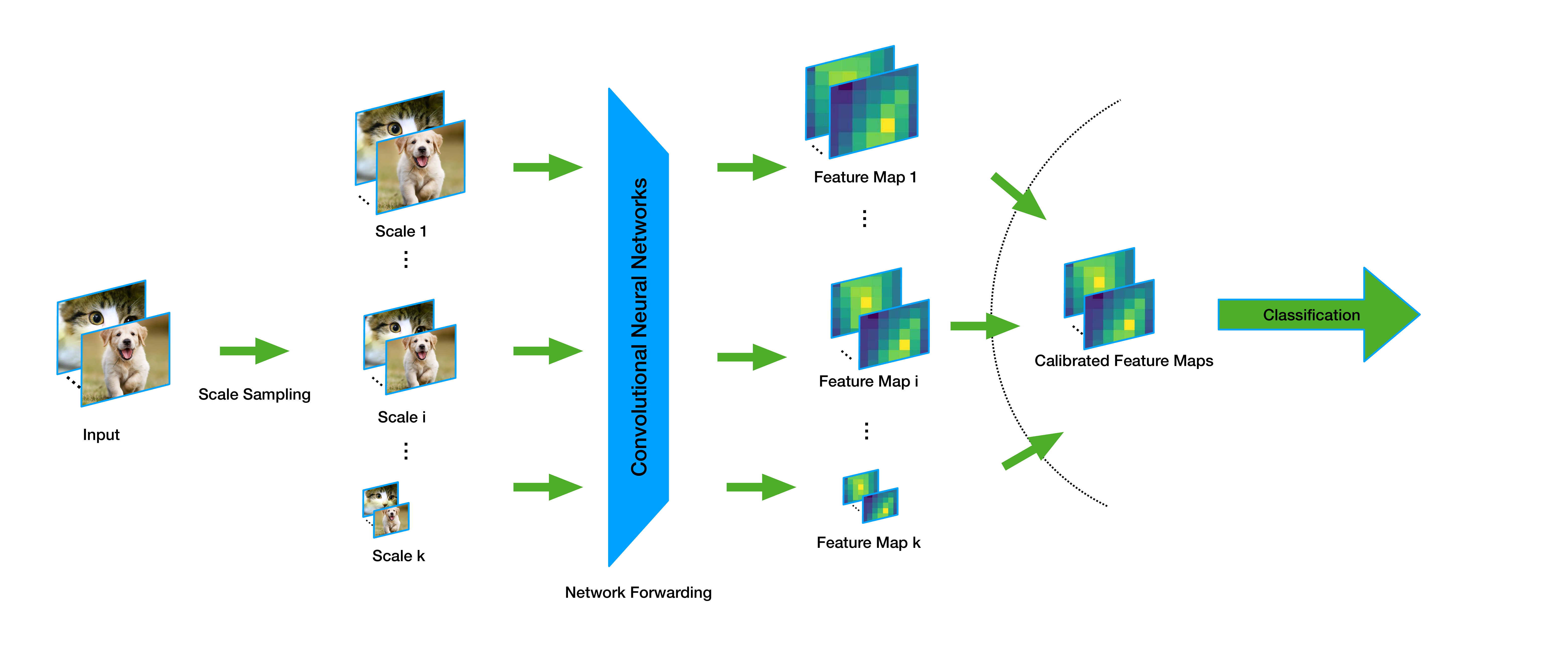}
\vspace*{-0.5in}

\caption{The overall pipeline of training networks for classification problems: the figure above shows the training step. During the sampling phase, fix-sized input images are transformed to samples with different spatial sizes. Samples are forwarded through a convolutional neural network and spatially calibrated before presented to the final linear classifier.}
\label{fig: pipeline}
\vspace{-0.15in}
\end{figure*}
\end{enumerate}

%% file: related.tex
\section{Related Work}
\textbf{Training Regime}. Many classic convolutional neural network models have much redundancy of channels and many changes in training regime have been proposed to better utilize network parameters and improve generalizations. Dropout\cite{srivastava2014dropout} is another approach that helps regularization by deactivating neurons of networks with some probability. Following research work in dropping layers during training\cite{wan2013regularization, wu2015towards}, RePr\cite{prakash2019repr} temporarily drops channels from networks and restore them with new initialization. All these work involve direct manipulation to networks during training. On the other hand, Mixup\cite{zhang2017mixup} is a data-driven approach that trains networks on convex combination of training samples. Some other simple yet efficient approaches in convolutional neural network training is described in \cite{he2019bag}. We shall see in section \ref{sec:method} that our approach works in a similar fashion to Mixup without directly manipulating network layers. 
\medskip 

 \textbf{Size Discrepancy between Training and Testing}. Previous work have discovered that the size discrepancy between training and testing decreases accuracy in testing. Few attempts tried to tackle this problem. Progressive resizing\cite{karras2017progressive} proposed to progressively increase sizes of images presented to generator and discriminator in GANs\cite{goodfellow2014generative}. FixRes\cite{touvron2019fixing} identifies the limitation of current data augmentation strategies that implicitly causes discrepancy between the actual sizes of objects in train and test phases. Their approach worked on a joint optimization approach and achieved accuracy advancement when training with identical and smaller image sizes. We shall see in experiment section that FixRes does a good job on classifying high solution images but benefits of inferecing with low resolution images are not fully exploited. 
 
\medskip
 \textbf{Normalization}. Normalization methods stablizes the internal activations of networks. Batch normalization (BN)\cite{ioffe2015batch} is a widely used ingredient that normalize internal activations with statistics of training batches. Several variants of BN are proposed. Layer normalization \cite{ba2016layer} normalizes activations using channel-wise statistics. Instance Normalization\cite{ulyanov2016instance}, on the other hand, normalizes each sample independently. Group normalization \cite{wu2018group}, which divides channels into several groups and normalizes each group independently, stabilizes training procedure for tasks where small batches are incorporated. Our analysis in the experiment section identifies the fact that these normalization schemes do not necessarily improve performance of networks trained with SCT. Further, we propose our novel normalization method in response to the problem.


%% file: method.tex
\DeclareFixedFont{\ttb}{T1}{txtt}{bx}{n}{12} 
\DeclareFixedFont{\ttm}{T1}{txtt}{m}{n}{12}  

\section{Method}
\label{sec:method}

In this section, we clarify the insight of our SST, describe its key components, and detail its implementation.
\subsection{Motivation}

There is no general agreement on what sizes of images should be chosen for training and testing. High resolution images expose more semantic information whereas low resolution images allow faster training and inference. In practice, the tradeoff of "accuracy vs speed"\cite{tan2019efficientnet} is ubiquitous. Simply feeding low resolution images to networks trained with high resolution images results in a catastrophic accuracy drop as shown in Figure \ref{fig: de}. One trivial solution is to store networks trained on various sizes of images and select a proper one for actual demand. However, this solution has two main drawbacks: it introduces additional maintenance cost to store separate models and training with low resolution image usually leads to low accuracy as shown in Table \ref{table:craw}. 

Then, a natural and greedy question to ask is: \textit{is it possible to get a \textbf{single} network that provides decent accuracy no matter what sizes of images are used for testing?}

\begin{table}[htbp]
	\centering
	\begin{tabular}[b]{ lllll }
		\toprule
		 Train \& Test Size& {64} & {128} & {224} & {384} \\
		 \midrule
		 Accuracy(\%) & 63.4 & 73.1 & 77.1 & 78.3 \\
		 \midrule
	\end{tabular}
	 \caption{Results of ResNet-50 with different sizes of images on ImageNet.}
    \label{table:craw}
\end{table}

In this paper, we present Scale Calibrated Training(SCT) to give a positive answer to the aforementioned question. SCT as a training regime can be applied to any CNN models and make networks more robust in response to variance in spatial size inputs.

\subsection{Scale Calibrated Training}
We demonstrate the process of \textit{Scale Calibrated Training(SCT)}. The pipeline of our method is presented in Figure \ref{fig: pipeline} and the core components are as follows.

\medskip 

\textbf{Scale Sampling Phase}. In this phase, a batch of raw images is first passed through standard augmentation procedure, which is usually composed of operations such as \textit{ResizedCrop} and \textit{RandomHorizontalFlip} followed by a normalization operation. Then, the transformed batch data, denoted by $\textit{X}$, of dimension $B \times C \times H \times W$ is passed to the \textbf{scale sampling layer}. $\textit{X}$ is then transformed to $M$ batches through bi-linear interpolation with several pre-defined and well-spaced scales\footnote{Generally, it is a good practice to make the interval between scales as the downsample factor from the input of output in most networks. We follow this design in all of our experiments.} $\{s_1, s_2, ..., s_M\}$. To this end, the output of the sampling layer is of dimension $\{B \times C \times H^i \times W^i\}_{i=1}^M$, where $M$ is the number of different scales included in the run of \textit{SCT}. In other words, the sampling layer generates M batches of data each of a distinct scale and we call it the scale collection of mini-batch $\textit{X}$.

\medskip
\textbf{Forwarding and Calibrating Phase}. Components of the scale collection of mini-batch $\textit{X}$ are then forwarded to the deep neural networks simultaneously. There is no requirement on DNN models and any types of models can be used in this phase. Finally, since spatial sizes are different cross scales, each with dimension $\{B \times C_{out} \times H^i_{out} \times W^i_{out}\}_{i=1}^M$, the final features are generated through adaptive average pooling so that features from different scales can be calibrated to the same dimension, which makes it plausible to use linear classifiers with shared weights to classify images. 

\medskip
\textbf{Loss and Backwards}. With standard stochastic gradient descent, in each iteration, the loss is calculated over a mini-batch of data and then propagated backward to update model parameters. In \textit{SCT}, the loss is calculated over the scale collection of a mini-batch generated by the scale sampling layer. We further assign tunable weights to each scale, the denoted as $\alpha_i$, which refers to weights of i-th scale in loss computation. The general formula of loss calculation with mini-batch $\textit{X}$ is described as follows: 
\begin{equation}
    \label{total loss}
    \mathcal{L} = \sum_{i=1}^{M} \alpha_i \frac{1}{B}\sum_{j=1}^{B} L_{cls}(f(x_j^i), y_j^i) 
\end{equation}
where $f(\cdot)$ denotes the sequential execution of deep neural networks and $x_j^i$, $y_j^i$ denotes sample image $j$ at i-th scale and its corresponding label respectively, $B$ denotes batch size and $M$ denotes the total number of scales included in \textit{SCT}. Gradients can be computed with the loss as normal with back propagation and weights are updated correspondingly.  

\medskip
\textbf{Testing Phase}. Testing a model trained by \textit{SCT} does not introduce any extra cost. Testing images with different scales can be directly forwarded by skipping the scale sampling phase. The forwarding phase in testing works exactly the same as that in training. 

\subsection{Refined Design of Batch Normalization}
In general, vanilla batch normalization normalizes entries in feature maps of batch $X$ as follows:
\begin{equation}
    \hat{x} = \gamma\frac{x - E[X]}{\sqrt{Var[X] + \epsilon}} + \beta
\end{equation}
where $\gamma$ and $\beta$ are scaling and shifting factors and $\epsilon$ is the smoothing constant. In training phase, $\mu$ and $\sigma$ are calculate over mini-batch $B$ and in testing phase, those values are directly extracted from running statistics. 

We describe \textit{Scale-Specific Batch Normalization(S-BN)} in replacement of vanilla batch normalization layer in \textit{SCT}.

\textbf{Scale-Specific Batch Normalization(S-BN).} Batches in the same scale collection are passed through regular convolution layers and batches at each scale are normalized by a specific BN layer. In other words, \textit{S-BN} makes parameters of batch normalization layers non-shareable between different spatial scales of inputs. Meanwhile, \textit{S-BN} computes batch statistics at each specific scale and uses scale-specific scaling and shifting parameters($\gamma$ and $\beta$ in BN). The feature map of an image sample \textit{x} at scale \textit{i} is normalized by:
\begin{equation}
    \hat{x^i} = \gamma^i\frac{x^i - E[X^i]}{\sqrt{Var[X^i] + \epsilon}} + \beta^i
\end{equation}
where $X^i$ is the batch of samples at scale $i$ to which $x^i$ belongs.  The running mean and variance are updated by batch statistics at scale $i$ correspondingly. Therefore, we have $M$ disjoint set of parameters $\{\gamma^i, \beta^i, \mu^i, \sigma^i\}_{i=1}^M$ for each \textit{S-BN} layer where $M$ is total number of scales.

Inference with \textit{S-BN} does not introduce additional costs: parameters of BN layers are switched corresponding to the input scale on-the-fly in the forward pass of inference. In experiment section, We show that \textit{S-BN} significantly improves test accuracy with all training scales over the vanilla batch normalization and we investigate the reasons behind the performance in the experiment section as well. 

%% file: experiments.tex
\section{Experiments}
We evaluate the performance of our method SCT with Scale-Specific Batch Normalization(S-BN) on image classification benchmarks. To better illustrate the effectiveness of our method, we simply set uniform weights of $\frac{1}{M}$ where $M$ is the total number of scales to each scale in the calculation of loss. 

\subsection{ImageNet Image Classification}
\label{sec:imgc}
We conduct exhaustive experiments on ImageNet image classification task to evaluate the effectiveness of SCT. We demonstrate the improvement of SCT when testing images are in the same size as training images and when testing images are in low resolution. We use standard ResNet-18 and ResNet-50 with batch norm trained on images of $224 \times 224$ as baselines. All models are trained with batch size of 256 on 8 Nvidia Titan XP GPUs over 120 epochs and learning rates are annealed from 0.1 to 0 with a cosine scheduler\cite{xie2018bag}. For SCT, we use three different schemes: the \textbf{standard} setting with only scale of 224, \textbf{SCT-A} including scales of 320, 224, 192, and \textbf{SCT-B} including scales of 320, 224, 192, 128, 64. 

\subsubsection{Testing with Standard Training Resolution}
Table \ref{table:imagnet} illustrates the validation accuracy on ImageNet evaluated at the standard training scale $224 \times 224$. Obviously, all models trained with SCT have acquired performance gain to some extent. No matter what normalization schemes used, training using SCT provides a better accuracy over the baseline model while networks equipped with GN can outperform models with BN. It is also remarkable that using Scale-Specific BN with SCT can further improve accuracy of networks where ResNet-18 and ResNet-50 have gained 2.1\% and 1.7\% accuracy bonus respectively. 
\begin{table}[]
	\centering
	\vspace{-5mm}
	\begin{tabular}[b]{lllll}
		\toprule
		\small
		 {\textbf{Model}} & {-} & {\textbf{Scheme}} & {\textbf{Top-1(\%)}} & {\textbf{Top-5(\%)}} \\
		 \midrule
		 \multirow{8}{*}{ResNet-18} 
		     & \multirow{3}{*}{BN} &Standard &71.1 & 90.0\\
		                      & &SCT-A  & 71.9 & 90.5\\
		                      & &SCT-B & 71.3 &90.2\\ \cline{2-5}
		     & \multirow{3}{*}{GN} &Standard &69.9 &89.3 \\
		                      & &SCT-A &71.8 &90.4 \\
		                      & &SCT-B
		                        &72.0  &90.5 \\ \cline{2-5}
		     & \multirow{2}{*}{S-BN} &SCT-A &72.8 {\color{red}(+1.7)} &91.0\\
		                  &&SCT-B&73.2 {\color{red}(+2.1)} &91.2{\color{red}(+1.2)} \\
		  \toprule
		  \multirow{8}{*}{ResNet-50}
		     & \multirow{3}{*}{BN} &Standard &77.1 &93.5 \\
		                  & &SCT-A &77.3&93.3\\ \cline{2-5}
		     & \multirow{3}{*}{GN} &Standard &76.9 &93.3 \\
		                       & &SCT-A &77.9&93.9 \\ 
		                           \cline{2-5}
		     & \multirow{2}{*}{S-BN} &SCT-A &78.3 {\color{red}(+1.4)}  &94.1 \\
		                      & &SCT-B &78.8 {\color{red}(+1.7)}  &94.5{\color{red}(+1.0)} \\
		  \bottomrule
	\end{tabular}
	\caption{Validation accuracy for ImageNet}
    \label{table:imagnet}
\end{table}

\subsubsection{Testing with Low Resolution Images}
\label{sec: res-low}
One of the key motivation of this work is to improve test accuracy of low resolution images because doing so can reduce inference time without modifying network structures. We showcase our experiment results of ResNet trained by SCT with S-BN when testing with low resolution images in this section. 

Table \ref{table:lowres} shows test results of SCT training in response to low resolution images. The salient fact is that the baseline model(standard ResNet-18 trained without SCT) does unsurprisingly poorly on low resolution images. When SCT is invoked, the accuracy of testing with size of $192 \times 192$ and size of $128 \times 128$ have improved 4.3\% and 11.5\% respectively. 

\begin{table}
\centering
\footnotesize
	\begin{tabular}{lll llll }
		\toprule
		 {\textbf{Model}} & {\textbf{Method}} & {\textbf{Scheme}} & {\textbf{Top-1(\%)@192}} & {\textbf{Top-1(\%)@128}} \\
		 \midrule
		  \multirow{8}{*}{ResNet-18}
		     & \multirow{2}{*}{BN} &Standard &67.6&55.3 \\
		                  & &SCT-A&71.1&65.0\\ 
		                  \cline{3-5}
		     & \multirow{2}{*}{GN} 
		                       &SCT-A&70.1&64.1 \\
		                       & &SCT-B&70.6&65.3 \\
		                           \cline{3-5}
		     & \multirow{2}{*}{S-BN} &SCT-A &71.4 &65.8 \\
		                  & &SCT-B & 71.9 {\color{red}(+4.3)} & 66.8 {\color{red}(+11.5)} \\
		  \bottomrule
	\end{tabular}
	\caption{Low resolution validation accuracy for ImageNet}
    \label{table:lowres}
\end{table}
This section of experiments demonstrates that single network trained with SCT can perform reasonably well with various testing sizes, which saves extra cost on storing multiple models for different resolutions.

\subsubsection{SCT with Limited Capacity Models}

Unlike over-parameterized convolutional neural networks, which often have much redundancy, efficient deep learning models\cite{tan2019efficientnet, sandler2018mobilenetv2} have limited capacity, which makes the accuracy improvement more challenging. Low-bit quantization is widely used in practice, which also lowers the model compacity. 

In this section, we present the efficiency of SCT in lightweight network MobileNetV2\cite{sandler2018mobilenetv2} and ResNet-18 with low-bit quantization. Specifically, the quantization training follows PACT\cite{choi2018pact} which finetunes the float-point pretrained model 20 epochs. 

Detailed results are shown in Table \ref{table:mobile}, where the SCT with S-BN provides steady accuracy increase for both MobileNet-V2 and 8-bit ResNet-18. The result in this section shows that our method can also work with networks with limited capacity.

\begin{table}[H]
	\centering
	\setlength{\belowcaptionskip}{-0.2cm}
	\footnotesize
	
	\begin{tabular}[b]{ lllll }
		\toprule
		 {\textbf{Model}} & {\textbf{Method}} & {\textbf{Scheme}} & {\textbf{Top-1(\%)}}  & {\textbf{Top-5(\%)}} \\
		 \midrule
		 \multirow{3}{*}{MobileNetV2} 
		     & \multirow{2}{*}{BN} &Standard &72.1  &90.5 \\
	
		                      & &SCT-B& 72.3 &90.6\\ 
		      \cline{3-5}\\
		     & \multirow{1}{*}{S-BN} &SCT-B &73.3 {\color{red}(+1.2)} &91.1 {\color{red}(+0.8)}\\
		       \cline{3-5} \\
		  \multirow{3}{*}{ResNet-18(\textbf{8 bit})}
		     & \multirow{1}{*}{BN} &Standard &71.3 &90.1 \\
		      \cline{3-5} \\
		     & \multirow{1}{*}{S-BN} &SCT-B &73.1 {\color{red}(+1.8)} &91.1 {\color{red}(+1.0)} \\
		  \bottomrule
	\end{tabular}
	\caption{Validation accuracy for efficient deep model}
    \label{table:mobile}
\end{table}
Our results in Section \ref{sec:imgc} demonstrates the universal effectiveness of our method. In summary, our method is compatible with different types of deep neural networks and improves the accuracy of \textbf{a single model} when evaluated with different resolutions.

\subsection{Ablation Study}
We present our ablation study on different normalization approaches in \textit{SCT} with deeper analysis. Besides vanilla batch normalization\cite{ioffe2015batch} and our refined designs of BN, we further include group normalization\cite{wu2018group}, fix-up initialization\cite{zhang2019fixup}. 

\textbf{Why Vanilla BN Fails.} We first present our analysis on the failure case of vanilla BN designs in \textit{SCT} by showing results on CIFAR-10\cite{krizhevsky2014cifar} benchmark dataset. We follow the original ResNet\cite{he2016deep} to build ResNet-32 and ResNet-110 with Basic Blocks. We include two sets of scale combinations in \textit{SCT}: $\{32 \times 32, 24 \times 24\}$ and $\{32 \times 32, 16 \times 16\}$ where $32 \times 32$ is the standard scale in CIFAR-10 training and testing. We train networks over 160 epochs with a batch size of 128 on a single NVIDIA 2080 GPU. The initial learning rate is 0.1, decayed by a factor of 10 at epoch 80 and 120 respectively. All experiments are run 3 times and we report the averaged accuracy. 

\begin{figure*}[htp]
\centering
\subfloat[Validation Results of Resnet-32 on Cifar-10]{
  \includegraphics[width=\columnwidth]{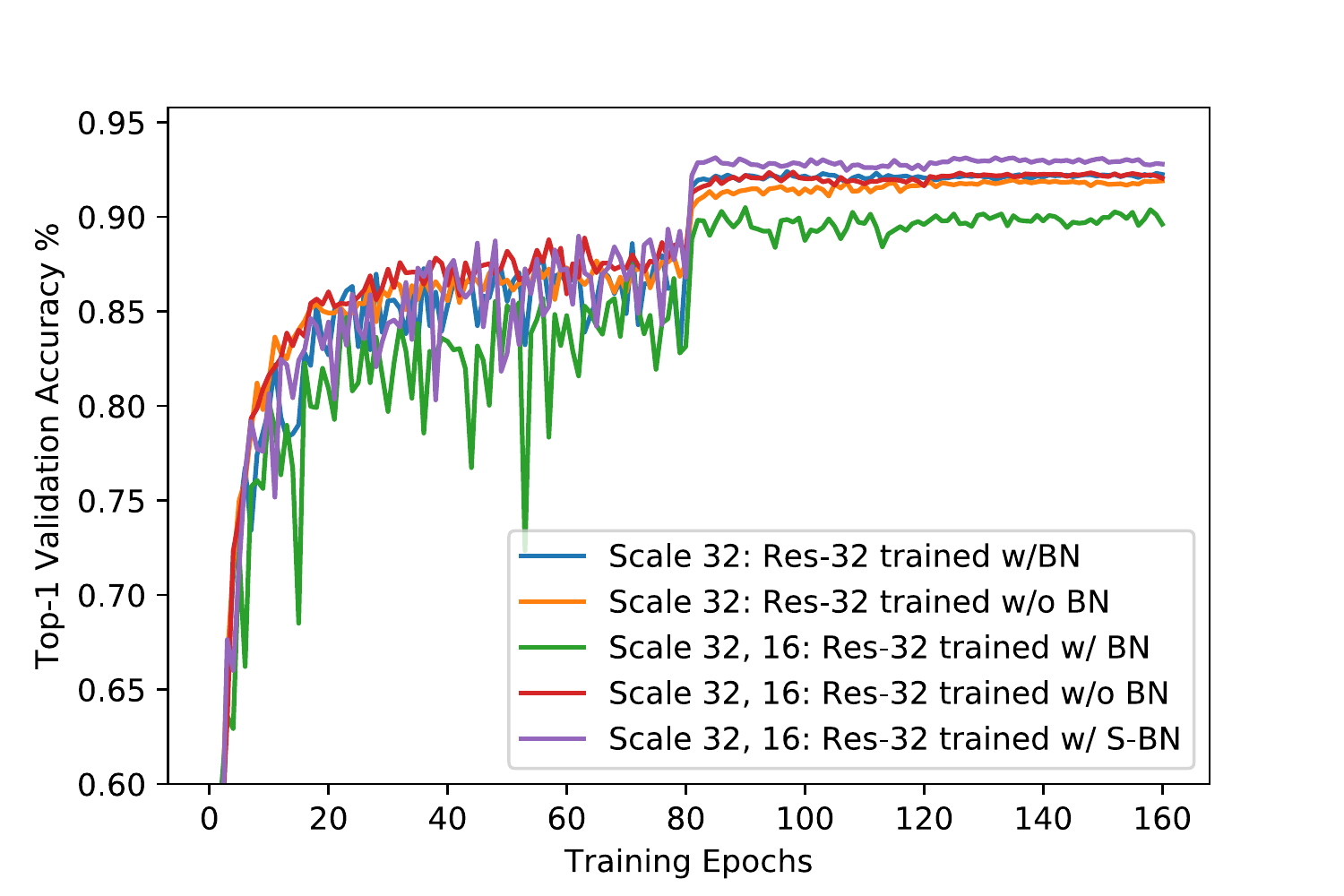}
  \label{fig:resnet-32}
}
\subfloat[Validation Results of Resnet-110 on Cifar-10]{
  \includegraphics[width=\columnwidth]{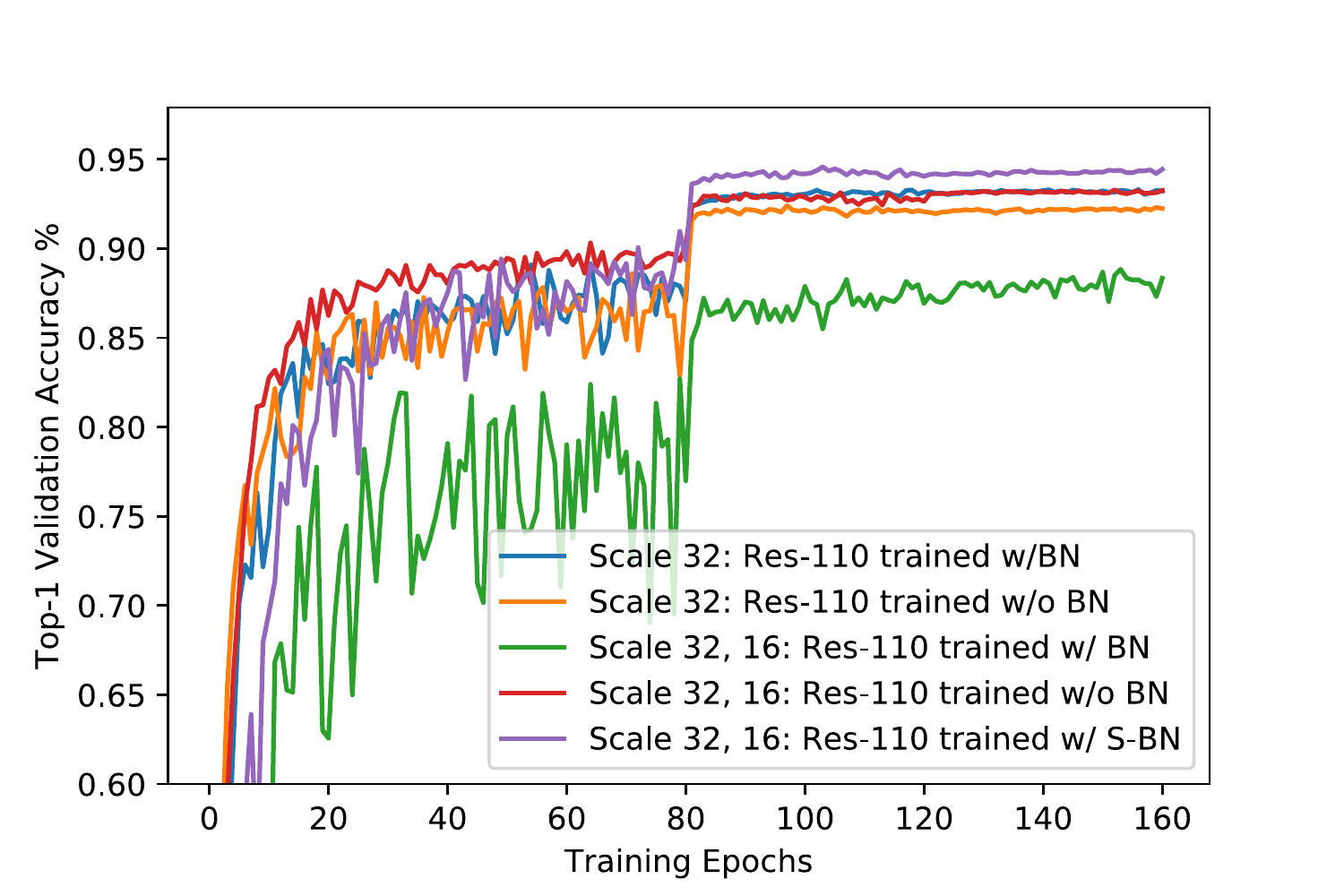}
  \label{fig:resnet-110}
}
\caption{ResNet Validation Accuracy of SCT: ResNets with batch normalization show an obvious accuracy drop when scales of input have dramatic distinction. ResNets without batch normalization do not suffer from this issue and benefits from SCT. Networks with exclusive batch normalization trained with SCT outperform all other settings. }
\vspace{-5mm}
\label{fig-cifar}
\end{figure*}

As shown in Table \ref{table:cifar} and Figure \ref{fig-cifar}, when scale difference  in \textit{SCT} is not significant, accuracy of all models are improved. However, when this difference becomes considerable, models with vanilla batch norm suffer from an accuracy loss. Furthermore, ResNet-110 with BN loses more accuracy than ResNet-32 does(\textbf{-7.34} and \textbf{-1.75}). 

Networks without vanilla Batch Normalization does not reveal similar behavior: Group Normalization is more robust in response to scale variation of inputs and we speculate this is because it doe not track running statistics and training with both combinations of scales increases the accuracy. Similarly, networks with Fix-up initialization also shows steady increase when multi-scale images are included. However, both networks with both GN and Fix-up cannot achieve significantly higher accuracy than networks trained with single-scale and vanilla BN.

On the other hand, training with \textit{S-BN} recovers networks from accuracy loss when variance in scales becomes large and significantly outperforms standard training strategy with vanilla BN.

To further demonstrate the issue of vanilla BN, we analyze the distribution of intermediate feature maps of network layers. We provide the visualization of feature map distribution at the inference phase. In other words, we visualize the distribution of intermediate activation of ResNet-32 built with batch normalization and Fixup initialization respectively from the last layer of each stage where stages refer to a collection of consecutive layers whose outputs share the same spatial size. 

\begin{table}[]
	\centering
	\vspace{2mm}
	\footnotesize
	\begin{tabular}[b]{ ccccc }
		\toprule
		 {\textbf{Model}} & {\textbf{Method}} & \multicolumn{3}{c}{\textbf{mean(\%) $\pm$ std(\%)}} \\
		\cline{3-5}
		  &  & 32 & 32, 28 & 32, 16 \\
		 \midrule
		 \multirow{4}{*}{ResNet-32} 
		     & BN &92.58(0.21) &93.12(0.19) &90.83(0.28) \\ \cline{2-5}
		     & GN & 90.55(0.17)&91.47(0.31) &91.22(0.28) \\ \cline{2-5}
		     & Fixup &91.92(0.20) & 92.63(0.25)&92.71(0.39) \\  \cline{2-5}
		     & S-BN & - &\textbf{93.35(0.23)} &\textbf{92.96(0.11)} \\
		  \toprule
		  \multirow{4}{*}{ResNet-110}
		     & BN & 93.59(0.13)&93.83(0.20) &86.25(0.33) \\ \cline{2-5}
		     & GN &89.25(0.19) & 92.0(0.22)& 91.01(0.25) \\ \cline{2-5}
		     & Fixup & 92.27(0.29)& 92.97(0.28)& 93.38(0.32)\\  \cline{2-5}
		     & S-BN & -&\textbf{94.28(0.19)} & \textbf{94.46(0.22)}\\

		  \bottomrule
	\end{tabular}
	\caption{Validation accuracy for CIFAR-10}
    \label{table:cifar}
\end{table}

\begin{figure*}[]
\centering
\includegraphics[width=2\columnwidth]{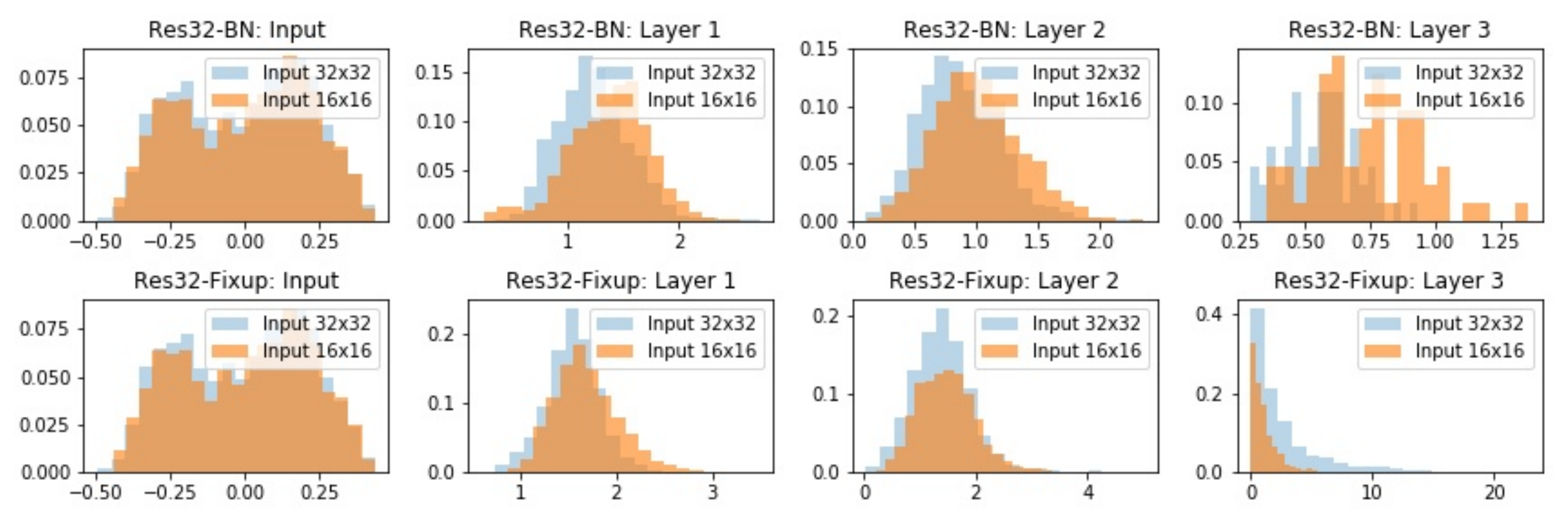}
\caption{Distribution of feature maps outputted by ResNet-32 with BN and Fixup initialization respectively}
\label{fig-stat}
\vspace{-5mm}
\end{figure*}

Figure \ref{fig-stat} provides a visualization for the aforementioned experiment. It can be seen from the plot that the initial distribution difference between images with different scales is marginal. However, as the network goes deeper, the network built with BN shows a distinct discrepancy between feature maps of two scales. When it comes to the last stage, the distribution of outputs corresponding to different input scales becomes divergent. In contrast, the distribution of feature maps outputted by the network with Fixup initialization shows consistency between scales even for activations of deep layers.

The visualization of feature maps provides validation of our experimental results and speculation. It clearly identifies the drawback of sharing parameters of batch normalization between scales in \textit{SCT}.

\subsection{Comparison with State-of-the-art Training Schemes}

We compare the performance of SCT to other state-of-the-art training schemes to illustrate the effectiveness of our method on improving generalization when evaluated with training resolution. For all experiments in this section, SCT is built up with Scale-Specific Batch Normalization(S-BN) and the set of scales is: {320, 224, 196, 128, 96}. We denote our method as SCT for simplicity.  
\subsubsection{Comparison with other Training Schemes}
We first describe several training schemes we include for comparison: 

\textbf{Mixup\cite{zhang2017mixup}} is a training scheme that trains a network on convex combination of pairs of training samples and their labels. It can be viewed as a data-agnostic augmentation approach that regularizes networks to favor linear behavior in-between training images. 

\textbf{Repr}\cite{prakash2019repr} uses an approach that temporarily pruning and then restore subsets of channels of convolutional neural networks repeatedly. Dropped channels are re-initialized and included in training again at the end of this process. 

\textbf{FixRes}\cite{touvron2019fixing} tackles the problem of size discrepancy in training and testing by an augmentation-styled approach. It not only performs operations on training images but also applies particular operations on testing images. We report the best accuracy achieved by FixRes when evaluating with testing size of $128 \times 128$.

\textbf{Knowledge Distillation\cite{hinton2015distilling}} Knowledge Distillation(KD) is a widely used co-training strategy, based on a teacher-student knowledge transfer framework in which outputs from a pre-trained teacher model are "distilled" and transfer to student models with less parameters. We present a new knowledge distillation method, scale knowledge distillation, which distills knowledges from models trained with specific resolution of images to networks designed for other resolution of images. In experiments, we use two pre-trained ResNets with input scale of 320x320 and 192x192 to help the training of models that take input of scale 224x224. 

\smallskip
We evaluate all these schemes and compare it with SCT. Table \ref{table:comp} illustrates the comparison of different training strategies on ImageNet\cite{krizhevsky2012imagenet}. All of these methods are evaluated on testing image sizes of $224 \times 224$. In both ResNet-18 and ResNet-50, SCT outperforms all other training schemes and advances the accuracy by 0.5\% and 1.1\% respectively. If regarded as a pure training regime, SCT is still able to outperform all other existing training regimes, which further shows the effectiveness of our method. 
\begin{table}[htbp]
\centering
\vspace{3mm}
\setlength{\belowcaptionskip}{-0.3cm}
\begin{tabular}{llcc}
\hline
\textbf{Model} & \textbf{Method} & \textbf{Top-1(\%)} & \textbf{Top-5(\%)} \\  \hline
\multirow{5}{*}{ResNet-18}    & Mixup\cite{zhang2017mixup}  &71.9&90.5             \\
                              &  RePr\cite{prakash2019repr} &  72.7   & -              \\
                             &  RePr\footnotemark &  71.6  &90.2    \\
                             & Scale KD         & 72.1    & 90.6  \\
                              & \textbf{SCT}(Ours) & \textbf{73.2} & \textbf{91.2} 
\\ \hline
\multirow{5}{*}{ResNet-50}   & Mixup\cite{zhang2017mixup} & 77.5 &93.6  \\
                             & RePr\cite{prakash2019repr}  & 77.6      \\
                             & FixRes\cite{touvron2019fixing} &77.6 &-    \\
                             & Scale KD & 77.4   &93.5\\
                             & \textbf{SCT}(Ours)    &\textbf{78.8} &\textbf{93.9}    \\
\hline
\end{tabular}
\caption{Comparison of test accuracy from various training scheme with testing size $224 \times 224$ on ImageNet.}
\label{table:comp}
\end{table}

\subsubsection{Speed up Inference with Low Resolution Images}

As aforementioned in section \ref{sec: res-low}, SCT enable networks to classify low resolution images more accurately. In this section, we further compare the effectiveness of SCT with FixRes\cite{touvron2019fixing} and standard settings. For FixRes, we report the best accuracy achieved at specified test sizes no matter what training sizes are included. 

Experimental results on ResNet-50 are listed in Table \ref{table:lowres-fixres}. With ResNet-50, SCT outperforms both standard setting and FixRes on resolution of 128. When evaluating with image size of 224, SCT achieves comparable accuracy to FixRes tested at size of 448, which indicates its potential of inference speed with \textbf{4 times} faster latency in theory. 

\begin{table}[H]
\centering
\begin{tabular}{llll}
\hline
\textbf{Model} & \textbf{Method@Test Scale} & \textbf{Top-1(\%)}& \\  \hline
ResNet-50        & Standard@128     &  73.3       \\
ResNet-50        & FixRes@128             &  73.4       \\
ResNet-50        & SCT@128            & \textbf{73.9}   \\ \hline
ResNet-50        & FixRes@224            & 77.1    \\
ResNet-50        & FixRes@384            & 78.2    \\
ResNet-50        & FixRes@\textbf{448}   &78.8 \\ \hline
ResNet-50        & SCT@\textbf{224}      & \textbf{78.8}    \\
\hline
\end{tabular}
\caption{Comparison of SCT and FixRes\cite{touvron2019fixing} on low resolution images}
\label{table:lowres-fixres}
\end{table}

Experimental results in this section demonstrate that our method can effectively improve the accuracy of networks when evaluated at the standard training scales. Our method generally outperforms other effective training regimes achieving state-of-the-art accuracy for convolutional neural networks.

%% file: conclusion.tex
\section{Conclusion}
In this paper, we present Scale Calibrated Training(SCT), along with our novel Scale-Specific Batch Normalization module, as an effective training paradigm that significantly improves accuracy of the image classification tasks. Our method enables a single networks to be executable at multiple scales in inference and remarkably improves the accuracy over other existing state-of-the-art training regimes when evaluated with different image resolutions. We also conduct a detailed ablation study to identify the weakness of vanilla batch normalization in our method SCT and explain the advantage of our refined batch normalization design.